\documentclass[runningheads]{llncs}
\usepackage{fixme}
\usepackage{times}
\usepackage{xspace}
\usepackage{xcolor}
\usepackage{hyperref}
\usepackage{amssymb}
\usepackage{graphicx}
\usepackage{booktabs} 
\usepackage{enumitem}
\usepackage{multirow}
\usepackage{siunitx} 
\usepackage{subcaption}
\usepackage{array}

\fxsetup{ status=draft, theme=color, inline, margin=False}
\FXRegisterAuthor{shang}{ashang}{\color{red}S}
\FXRegisterAuthor{polo}{apolo}{\color{red}P}
\FXRegisterAuthor{cory}{acory}{\color{red}C}
\FXRegisterAuthor{jason}{ajason}{\color{red}J}

\usepackage{amsmath}
\DeclareMathOperator*{\argmin}{arg\,min}

\newcommand{\method}{\textit{ShapeShifter}\xspace}

\begin{document}

\title{ShapeShifter: Robust Physical Adversarial Attack on \\ Faster R-CNN Object Detector}
\author{
Shang-Tse Chen\inst{1} \and
Cory Cornelius\inst{2} \and
Jason Martin\inst{2} \and
Duen Horng (Polo) Chau\inst{1}}

\authorrunning{S.-T. Chen et al.}

\institute{Georgia Institute of Technology, Atlanta, GA, USA\\
\email{\{schen351, polo\}@gatech.edu}
\and
Intel Corporation, Hillsboro, OR, USA\\
\email{\{cory.cornelius, jason.martin \}@intel.com}
}

\maketitle

\begin{abstract}
Given the ability to directly manipulate image pixels in the digital input space,
  an adversary can easily generate imperceptible perturbations to fool a Deep Neural Network (DNN) image classifier,
 as demonstrated in  prior work.
In this work, we propose \method, an attack that tackles the more challenging problem of crafting physical adversarial perturbations to fool image-based object detectors like Faster R-CNN.
Attacking an object detector is more difficult than attacking an image classifier, as it needs to mislead the classification results in multiple bounding boxes with different scales.
Extending the digital attack to the physical world adds another layer of difficulty, because it requires the perturbation to be robust enough to survive real-world distortions due to different viewing distances and angles, lighting conditions, and camera limitations.
We show that the \textit{Expectation over Transformation} technique, which was originally proposed to enhance the robustness of adversarial perturbations in image classification, can be successfully adapted to the object detection setting.
\method can generate adversarially perturbed stop signs that are consistently mis-detected by Faster R-CNN as other objects, posing a potential threat to autonomous vehicles and other safety-critical computer vision systems.

\keywords{Adversarial attack \and Object detection \and Faster R-CNN}
\end{abstract}

\section{Introduction}

Adversarial examples are input instances that are intentionally designed to fool a machine learning model into producing a chosen prediction.
The success of Deep Neural Network (DNN) in computer vision does not exempt it from this threat.
It is possible to bring the accuracy of a state-of-the-art DNN image classifier down to near zero percent by adding imperceptible adversarial perturbations~\cite{Szegedy14,carlini2017towards}.
The existence of adversarial examples not only reveals intriguing theoretical properties of DNN, but also raises serious practical concerns on its deployment in security and safety critical systems.
Autonomous vehicle is an example application that cannot be fully trusted before guaranteeing the robustness to adversarial attacks.
The imperative need to understand the vulnerabilities of DNNs attracts tremendous interest among machine learning, computer vision, and security researchers.

Although many adversarial attack algorithms have been proposed, attacking a real-world computer vision system is difficult.
First of all, most of the existing attack algorithms only focus on the image classification task, yet in many real-world use cases there will be more than one object in an image. Object detection, which recognizes and localizes multiple objects in an image, is a more suitable model for many vision-based scenarios.
Attacking an object detector is more difficult than attacking an image classifier, as it needs to mislead the classification results in multiple bounding boxes with different scales~\cite{lu2017no}.

\begin{figure}[tb]
  \centering
  \includegraphics[width=\textwidth]{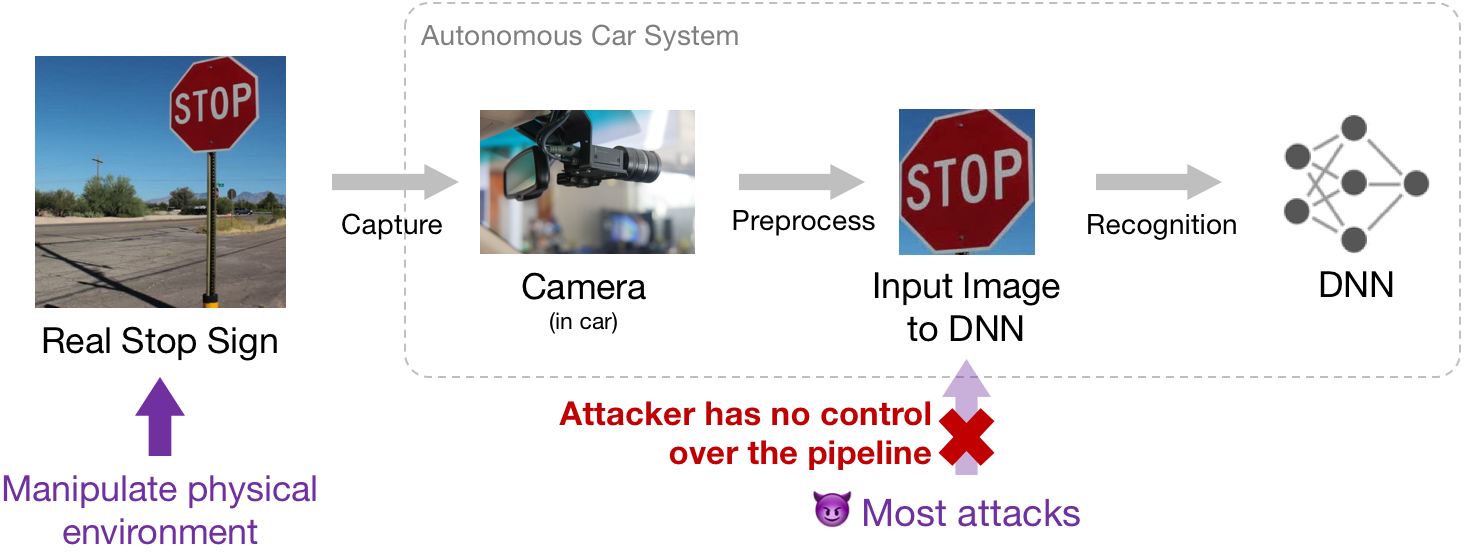}
  \caption{Illustration motivating the need of physical adversarial attack, from attackers' perspectives, as they typically do not have full control over the computer vision system pipeline.}
  \label{fig:physical}
\end{figure}

Further difficulty comes from the fact that DNN is usually only a component in the whole computer vision system pipeline.
For many applications, attackers usually do not have the ability to directly manipulate data inside the pipeline.
Instead, they can only manipulate the things outside of the system, i.e., those things in the physical environment.
\autoref{fig:physical} illustrates the intuition behind \textit{physical adversarial attacks}.
To be successful attacks, physical adversarial attacks must be robust enough to survive real-world distortions due to different viewing distances and angles, lighting conditions, and camera limitations.

There has been prior work that can either attack object detectors digitally~\cite{xie2017adversarial}, or attack image classifiers physically~\cite{kurakin2016adversarial,sharif2016accessorize,evtimov2017robust}. However, so far the existing attempts to physically attack object detectors remain unsatisfactory. A perturbed stop sign is shown in~\cite{lu2017adversarial} that cannot be detected by the Faster R-CNN object detector~\cite{ren2015faster}. However, the perturbation is very large and they tested it with poor texture contrast against the background, making the perturbed stop sign hard to see even by human. 
A recent short note~\cite{eykholt2017note} claims to be able to generate some adversarial stickers that, when attaching to a stop sign, can fool the YOLO object detector~\cite{redmon2017yolo9000} and can be transferable to also fool Faster R-CNN. However, they did not reveal the algorithm used to create the sticker and only show a video of indoor experiment with short distance\footnote{Three months after our work appeared on arXiv, the authors of~\cite{eykholt2017note} revealed their approaches in a separate paper~\cite{Song2018physical} and discussed the differences with our work.}. 
For other threat models of adversarial attacks in computer vision, we refer the readers to the survey of~\cite{akhtar2018threat}.

In this work, we propose \method, the first robust targeted attack that can fool a state-of-the-art Faster R-CNN object detector. To make the attack robust, we adopt the \textit{Expectation over Transformation} technique~\cite{athalye2017synthesizing,brown2017adversarial}, and adapt it from the image classification task to the object detection setting.
As a case study, we generate some adversarially perturbed stop signs that can consistently be mis-detected by Faster R-CNN as the target objects in real drive-by tests. Our contributions are summarized below.

  


\subsection{Our Contributions}

\begin{itemize}

\item To the best of knowledge, our work presents the first reproducible and robust targeted attack against Faster R-CNN~\cite{lu2017no}.
Recent attempts either can only do untargeted attack and requires perturbations with ``extreme patterns'' (in the researchers' words) to work consistently~\cite{lu2017adversarial}, or has not revealed the details of the method~\cite{eykholt2017note}.
We have open-sourced our code on GitHub\footnote{\url{https://github.com/shangtse/robust-physical-attack}}.

\item We show that the \textit{Expectation over Transformation} technique, originally proposed for image classification, can be applied in the object detection task and significantly enhance robustness of the resulting perturbation.

\item By carefully studying the Faster R-CNN object detector algorithm, we overcome 
non-differentiability in the model, 
and successfully perform optimization-based attacks using gradient descent and backpropogation.

\item We generate perturbed stop signs that can consistently fool Faster R-CNN in real drive-by tests (videos available on the GitHub repository), calling for imperative need to improve and fortify vision-based object detectors.




\end{itemize}

\section{Background}\label{sec:background}
This section provides background information of adversarial attacks and briefly describes the Faster R-CNN object detector that we try to attack in this work. 

\subsection{Adversarial Attack}
Given a trained machine learning model $C$ and a benign instance $x\in\mathcal{X}$ that is correctly classified by $C$, 
the goal of the untargeted adversarial attack is to find another instance $x'\in\mathcal{X}$, such that $C(x')\neq C(x)$ and $d(x,x')\leq \epsilon$ for some distance metric $d(\cdot, \cdot)$ and perturbation budget $\epsilon>0$. 
For targeted attack, we further require $C(x') = y'$ where $y'\neq C(x)$ is the target class.
Common distance metrics $d(\cdot, \cdot)$ in the computer vision domain are $\ell_2$ distance $d(x, x')=||x-x'||^2_2$ and $\ell_\infty$ distance $d(x, x')=||x-x'||_\infty$.

The work of \cite{Szegedy14} was the first to discover the existence of adversarial examples for DNNs.
Several subsequent works have improved the computational cost and made the perturbation highly imperceptible to human~\cite{goodfellow2014explaining,Moosavi16}. 
Most adversarial attack algorithms against DNNs assume that the model is differentiable, and use the gradient information of the model to tweak the input instance to achieve the desired model output~\cite{carlini2017towards}.
Sharif et al.~\cite{sharif2016accessorize} first demonstrated a physically realizable attack to fool a face recognition model by wearing an adversarially crafted pair of glasses.

\subsection{Faster R-CNN}
\label{sec:rcnn}
Faster R-CNN~\cite{ren2015faster} is one of the state-of-the-art general object detectors.
It adopts a 2-stage detection strategy. In the first state, a region proposal network is used to generate several class-agnostic bounding boxes called region proposals that may contain objects. In the second stage, a classifier and a regressor are used to output the classification results and refined bounding box coordinates for each region proposal, respectively.
The computation cost is significantly reduced by sharing the convolutional layers in the two stages.
Faster R-CNN is much harder to attack, as a single object can be covered by multiple region proposals of different sizes and aspect ratios, and one needs to mislead the classification results in all the region proposals to fool the detection.


\section{Threat Model}
Existing methods that generate adversarial examples typically yield imperceptible perturbations that fool a given machine learning model. 
Our work, following~\cite{sharif2016accessorize}, generates perturbations that are perceptible but constrained such that a human would not be easily fooled by such a perturbation. 
We examine this kind of perturbation in the context of object detection (e.g., stop sign).
We chose this use case because of object detector's possible uses in security-related and safety-related settings (e.g., autonomous vehicles).
For example, attacks on traffic sign recognition could cause a car to miss a stop sign or travel faster than legally allowed.

We assume the adversary has white-box level access to the machine learning model. 
This means the adversary has access to the model structure and weights to the degree that the adversary can both compute outputs (i.e., the forward pass) and gradients (i.e., the backward pass). 
It also means that the adversary does not have to construct a perturbation in real-time. 
Rather, the adversary can study the model and craft an attack for that model using methods like Carlini-Wagner attack~\cite{carlini2017towards}. 
This kind of adversary is distinguished from a black-box adversary who is defined as having no such access to the model architecture or weights. 
While our choice of adversary is the most powerful one, existing research has shown it is possible to construct imperceptible perturbations without white-box level access~\cite{Papernot17blackbox}. 
However, whether our method is capable of generating perceptible perturbations with only black-box access remains an open question. 
Results from Liu et al.~\cite{liu2017delving} suggest that iterative attacks (like ours) tend not to transfer as well as non-iterative attacks.

Unlike previous work, we restrict the adversary such that they cannot manipulate the digital values of pixels gathered from the camera that each use case uses to sense the world. 
This is an important distinction from existing imperceptible perturbation methods. 
Because those methods create imperceptible perturbations, there is a high likelihood such perturbations would not fool our use cases when physically realized. 
That is, when printed and then presented to the systems in our use cases, those perturbations would have to survive both the printing process and sensing pipeline in order to fool the system. 
This is not an insurmountable task as Kurakin et al.~\cite{kurakin2016adversarial} have constructed such imperceptible yet physically realizable adversarial perturbations for image classification systems.

Finally, we also restrict our adversary by limiting the shape of the perturbation the adversary can generate.
This is important distinction for our use cases because one could easily craft an odd-shaped ``stop sign'' that does not exist in the real world.
We also do not give the adversary the latitude of modifying all pixels in an image like Kurakin et al.~\cite{kurakin2016adversarial}, but rather restrict them to certain pixels that we believe are both inconspicuous and physically realistic.


\section{Attack Method}
\label{sec:method}
Our attack method, \method, is inspired by the iterative, change-of-variable attack described in~\cite{carlini2017towards} and the \textit{Expectation over Transformation} technique~\cite{athalye2017synthesizing,brown2017adversarial}. Both methods were originally proposed for the task of image classification.
We describe these two methods in the image classification setting before showing how to extend them to attack the Faster R-CNN object detector.

\subsection{Attacking an Image Classifier}

Let $F: [-1, 1]^{h\times w \times 3} \rightarrow \mathbb{R}^K$ be an image classifier that takes an image of height $h$ and width $w$ as input, and outputs a probability distribution over $K$ classes.
The goal of the attacker is to create an image $x'$ that looks like an object $x$ of class $y$, but will be classified as another target class $y'$.

\subsubsection{Change-of-variable Attack}
Denote $L_F(x, y) = L(F(x), y)$ as the loss function that calculates the distance between the model output $F(x)$ and the target label $y$. Given an original input image $x$ and a target class $y'$, the change-of-variable attack~\cite{carlini2017towards} propose the following optimization formulation.
\begin{equation}
\label{eq:cw_attack}
\argmin_{x'\in\mathbb{R}^{h\times w \times 3}}  L_F(\tanh(x'), y')  + c \cdot || \tanh(x') - x ||^2_2.
\end{equation}
The use of \textit{tanh} ensures that each pixel is between $[-1, 1]$. The constant $c$ controls the similarity between the modified object $x'$ and the original image $x$. In practice, $c$ can be determined by binary search~\cite{carlini2017towards}.

\subsubsection{Expectation over Transformation}

The \textit{Expectation over Transformation}~\cite{athalye2017synthesizing,brown2017adversarial} idea is simple: adding random distortions in each iteration of the optimization to make the resulting perturbation more robust.
Given a transformation $t$ that can be translation, rotation, and scaling, 
 $M_t(x_b, x_o)$ is an operation that transforms an object image $x_o$ using $t$ and then overlays it onto a background image $x_b$. $M_t(x_b, x_o)$ can also include a masking operation that only keeps a certain area of $x_o$. This will be helpful when one wants to restrict the shape of the perturbation.
 After incorporating the random distortions, equation (\ref{eq:cw_attack}) becomes
\begin{equation}
\label{eq:eot}
\argmin_{x'\in\mathbb{R}^{h\times w \times 3}} \mathbb{E}_{x\sim X, t\sim T}  \left[ L_F(M_t(x,  \tanh(x')), y') \right]  + c \cdot || \tanh(x') - x_o ||^2_2,
\end{equation}
where $X$ is the training set of background images.
When the model $F$ is differentiable, this optimization problem can be solved by gradient descent and back-propagation. The expectation can be approximated by the empirical mean.

\subsection{Extension to Attacking Faster R-CNN}
An object detector $F: [-1, 1]^{h\times w \times 3} \rightarrow (\mathbb{R}^{N\times K}, \mathbb{R}^{N\times 4})$
takes an image as input and outputs $N$ detected objects.
Each detection includes a probability distribution over $K$ pre-defined classification classes as well as the location of the detected object, represented by its $4$ coordinates. Note that it is possible for an object detector to output more or fewer detected objects, depending on the input image, but for simplicity we select top-$N$ detected objects ranked by confidence. 

As described in~\autoref{sec:rcnn}, Faster R-CNN adopts a 2-stage approach.
The region proposal network in the first stage outputs several region proposals, and the second stage classifier performs classification within each of the region proposals. Let $rpn(x) = \{r_1, \dots, r_m\}$, where each $r_i$ is a region proposal represented as its four coordinates, and let $x_r$ be a sub-image covered by region $r$.
Denote $L_{F_i}(x, y) = L(F(x_{r_i}), y)$ the loss of the classification in the $i$-th region proposal.
 We simultaneously attack all the classifications in each region proposal by the following optimization.

\begin{equation}
\label{eq:method}
\argmin_{x'\in\mathbb{R}^{h\times w \times 3}} \mathbb{E}_{x\sim X, t\sim T}  \left[ \frac{1}{m}\sum_{r_i\in rpn(M_t(x'))} L_{F_i}(M_t(x'), y') \right]  + c \cdot || \tanh(x') - x_o ||^2_2,
\end{equation}
where we abuse the notation $M_t(x') = M_t(x, \tanh(x'))$ for simplicity.
However, for computational issues, most models prune the region proposals by using heuristics like 
non-maximum suppression~\cite{ren2015faster}. The pruning operations are usually non-differentiable, making it hard to optimize equation~(\ref{eq:method}) end to end. Therefore, we approximately solve this optimization problem by
first run a forward pass of the region proposal network, and fixed the pruned region proposals as fixed constants to the second stage classification problem in each iteration. We empirically find this approximation sufficient to find a good solution.

\section{Evaluation}
We evaluate our method by fooling a pre-trained Faster R-CNN model with Inception-v2~\cite{szegedy2016rethinking} convolutional feature extraction component.
The model was trained on the Microsoft Common Objects in Context (MS-COCO) dataset~\cite{lin2014microsoft} and is publicly available in the Tensorflow Object Detection API~\cite{huang2017speed} model zoo repository\footnote{\footnotesize \url{http://download.tensorflow.org/models/object_detection/faster_rcnn_inception_v2_coco_2017_11_08.tar.gz}}. 

The MS-COCO dataset contains 80 general object classes ranging from people and animals to trucks and cars and other common objects.
Although our method can potentially be used to attack any classes,
we choose to focus on attacking the stop sign class due to its importance and relevance to self-driving cars, where a vision-based object detector may be used to help make decisions.
An additional benefit of choosing the stop sign is its flat shape that can easily be printed on a paper.
Other classes, like dogs, are less likely to be perceived as real objects by human when printed on a paper.
While 3D printing adversarial examples for image recognition is possible~\cite{athalye2017synthesizing}, we leave 3D-printed adversarial examples against object detectors as future work.

\subsection{Digitally Perturbed Stop Sign}
We generate adversarial stop signs by performing the optimization process described in \autoref{eq:method}.
The hyperparameter $c$ is crucial in determining the perturbation strength. A smaller value of $c$ will result in a more conspicuous perturbation, but the perturbation will also be more robust to real-world distortions when we do the physical attack later.

However, it is hard to choose an appropriate $c$ when naively using the $\ell_2$ distance to a real stop sign as regularization.
To obtain a robust enough perturbation, a very small $c$ needs to be used, which has the consequence of creating stop signs that are difficult for humans to recognize.
The $\ell_2$ distance is not a perfect metric for human perception, which tends to be more sensitive to color changes on lighter-colored objects.
Due to this observation, we only allow the perturbation to change the red part of the stop sign, leaving the white text intact.
This allows us to generate larger and more robust perturbation, while providing enough contrast between the lettering and red parts so that a human can easily recognize the perturbation as a stop sign.
The adversarial stop sign generated in~\cite{lu2017adversarial} does not consider this and is visually more conspicuous.
Automating this procedure for other objects we leave as future work.

We performed two targeted attacks and one untargeted attack. We choose person and sports ball as the two target classes because they are relatively similar in size and shape to stop signs.
Our method allows attackers to use any target classes, however the perturbation needs to achieve its means and fool the object detector.
For some target classes, this may mean creating perturbations so large in deviation that they may appear radically different from the victim class.
We also noticed that some classes are easier to be detected at small scales, such as \textit{kite}, while other classes (e.g., \textit{truck}) could not be detected when the object was too small.
This may be an artifact of the MS-COCO dataset that the object detector was trained on.
Nevertheless, ultimately the attacker has a choice in target class and, given ample time, can find the target class that best fools the object detector according to their means.

\begin{figure}[tb]
    \centering
     \begin{subfigure}[b]{0.25\textwidth}
        \includegraphics[width=\textwidth]{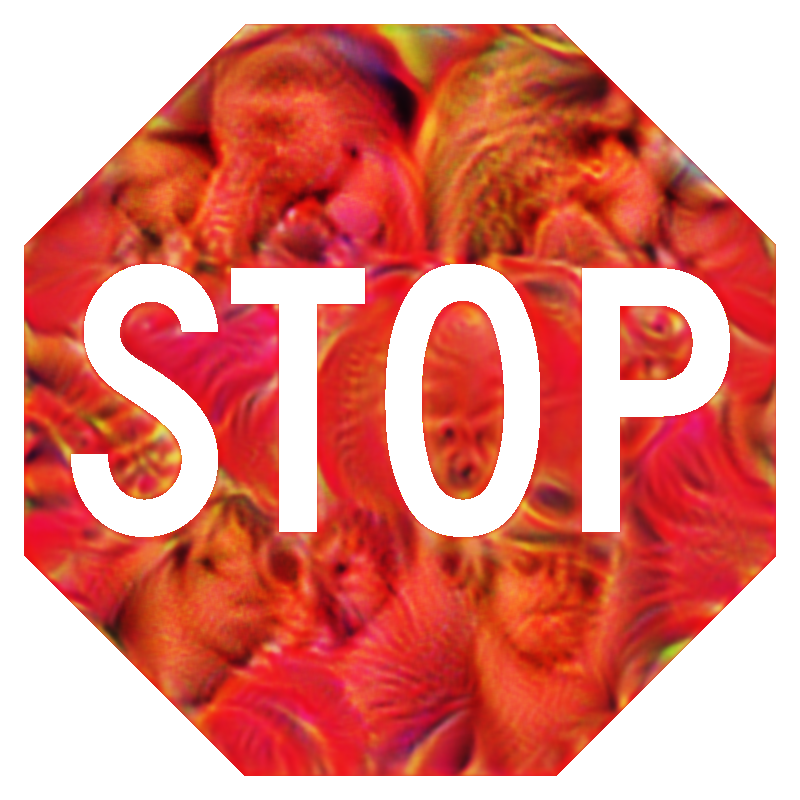}
        \caption{Person (low)}
         \label{fig:person_low}
     \end{subfigure}
     \hspace{0.01\textwidth}
    \begin{subfigure}[b]{0.25\textwidth}
        \includegraphics[width=\textwidth]{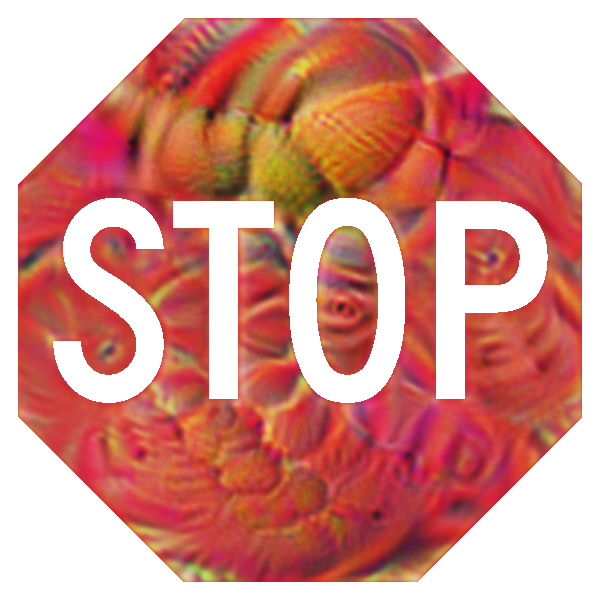}
        \caption{Sports ball (low)}
        \label{fig:ball_low}
     \end{subfigure}
     \hspace{0.01\textwidth}
     \begin{subfigure}[b]{0.25\textwidth}
        \includegraphics[width=\textwidth]{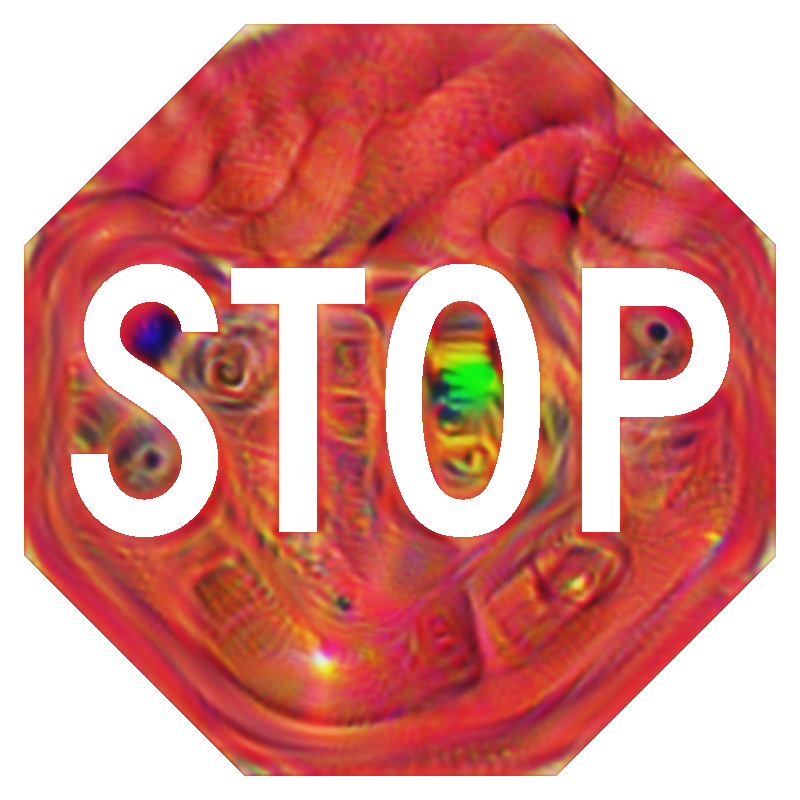}
        \caption{Untargeted (low)}
        \label{fig:untargeted_low}
     \end{subfigure}
\par\bigskip
    \begin{subfigure}[b]{0.25\textwidth}
        \includegraphics[width=\textwidth]{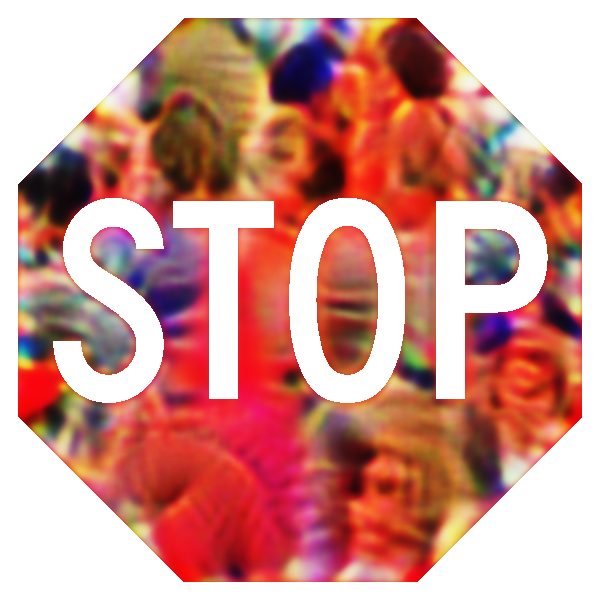}
        \caption{Person (high)}
         \label{fig:person_high}
     \end{subfigure}
     \hspace{0.01\textwidth}
    \begin{subfigure}[b]{0.25\textwidth}
        \includegraphics[width=\textwidth]{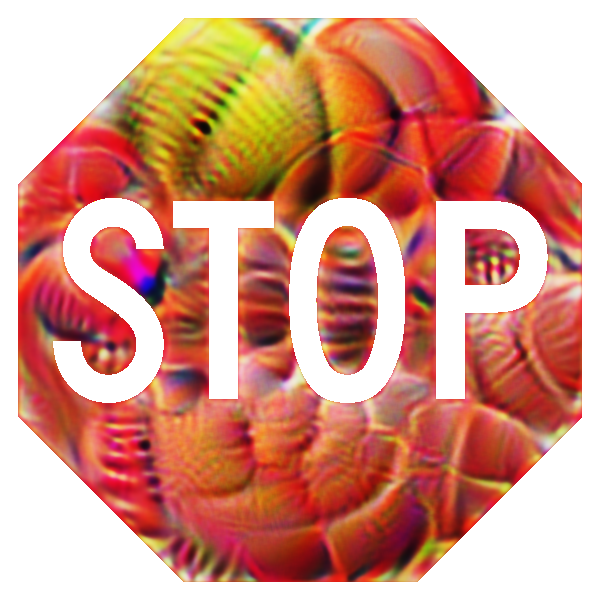}
        \caption{Sports ball (high)}
         \label{fig:ball_high}
     \end{subfigure}
     \hspace{0.01\textwidth}
    \begin{subfigure}[b]{0.25\textwidth}
        \includegraphics[width=\textwidth]{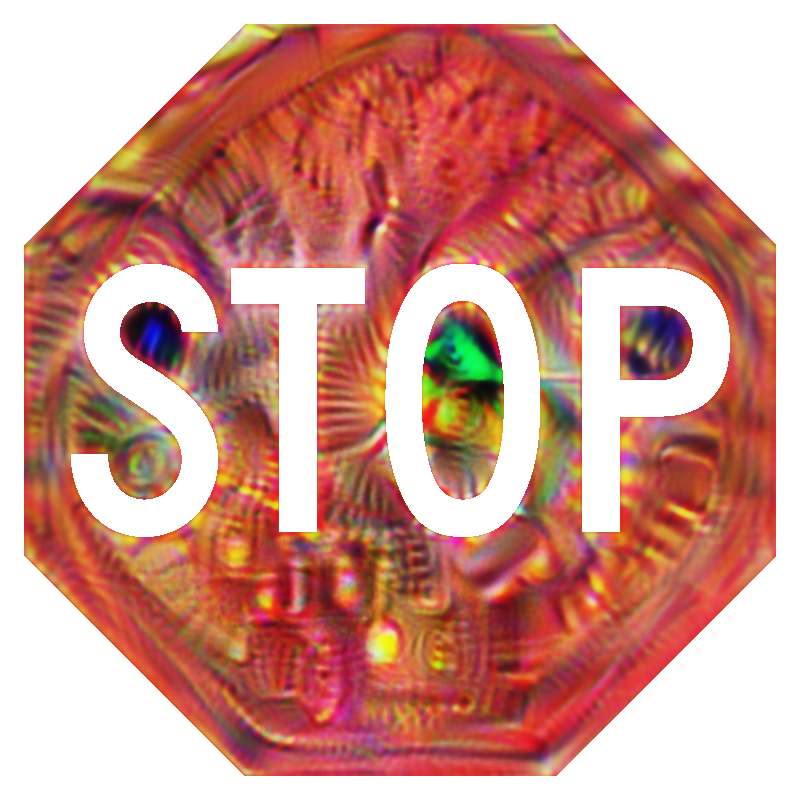}
        \caption{Untargeted (high)}
         \label{fig:untargeted_high}
     \end{subfigure}

          \caption{Digital perturbations we created using our method. Low confidence perturbations on the top and high confidence perturbations on the bottom.}
          \label{fig:digi_perturb}
\end{figure}

For each attack, we generated a high confidence perturbation and a low level perturbation.
The high confidence perturbations were generated using a smaller value of $c$, thus making them more conspicuous but also more robust.
Depending upon the target class, it may be difficult to generate an effective perturbation.
We manually chose $c$ for each target class so that the digital attack achieves high success rate while keeping the perturbation not too conspicuous,  i.e., we tried to keep the color as red as possible.
We used $c=0.002$ for the high confidence perturbations and $c=0.005$ for the low confidence perturbations in the ``sports ball'' targeted attack and the untargeted attack.
We used $c=0.005$ and  $c=0.01$ for the high and low confidence perturbations in the ``person'' targeted attack, respectively.
The 6 perturbations we created are shown in \autoref{fig:digi_perturb}.

\subsection{Physical Attack}
We performed physical attacks on the object detector by printing out the perturbed stop signs shown in~\autoref{fig:digi_perturb}.
We then took photos from a variety of distances and angles in a controlled indoor setting.
We also conducted drive-by tests by recording videos from a moving vehicle that approached the signs from a distance.
The lightning conditions varied from recording to recording depending upon the weather at the time.

 \subsubsection{Equipment}
 We used a Canon Pixma Pro-100 photo printer to print out signs with high-confidence perturbations, and an HP DesignJet to print out those with low-confidence perturbations\footnote{We used two printers to speed up our sign production, since a sign can take more than 30 minutes to produce.}.
 For static images, we used a Canon EOS Rebel T7i DSLR camera, equipped with a EF-S 18-55mm IS STM lens.
 The videos in our drive-by tests are shot using an iPhone 8 Plus mounted on the windshield of a car.

\begin{figure}[tb]
  \centering
  \includegraphics[width=\textwidth]{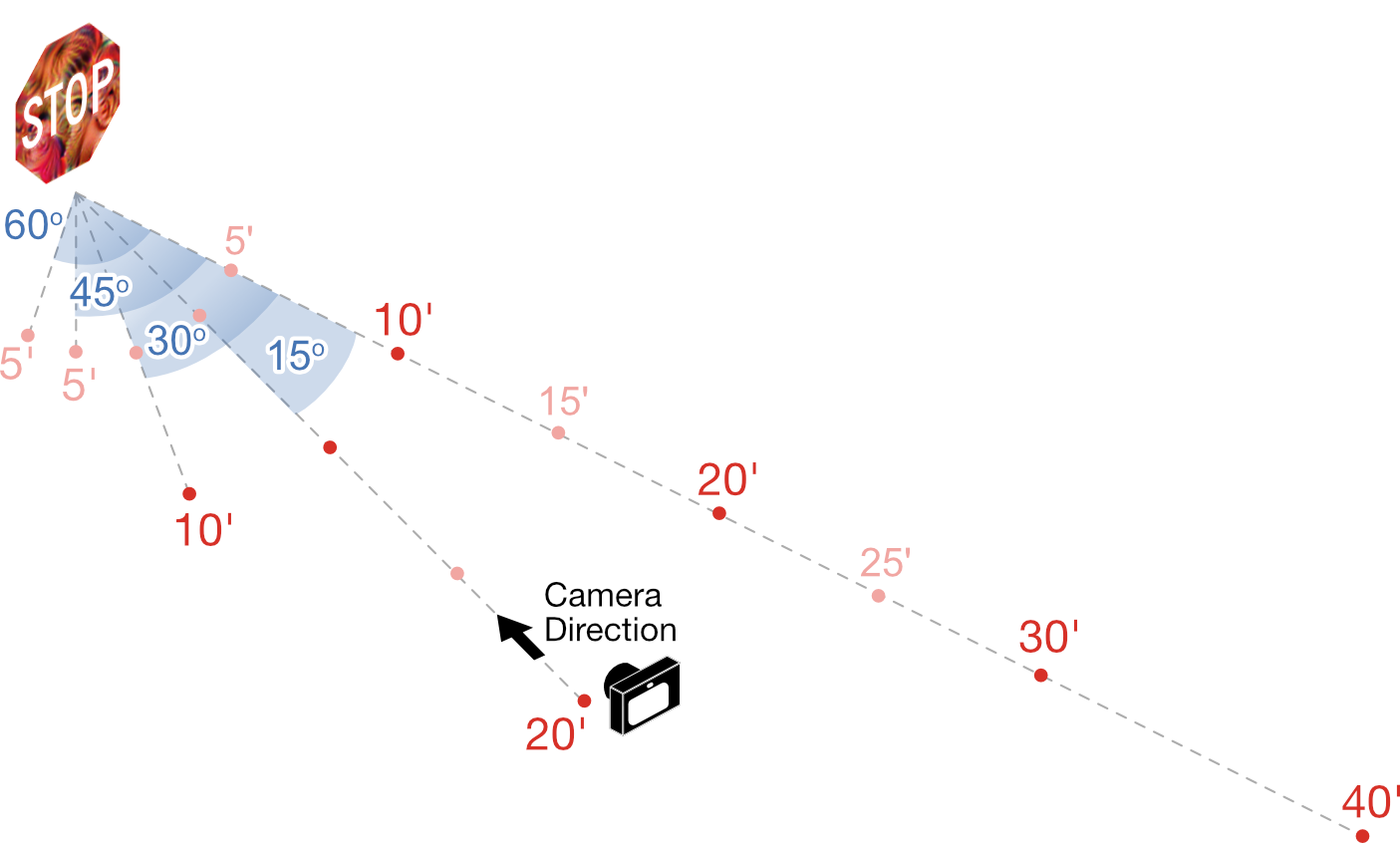}
  \caption{\textbf{Indoor experiment setup.} We take photos of the printed adversarial sign, from multiple  angles (\ang{0}, \ang{15}, \ang{30}, \ang{45}, \ang{60}, from the sign's tangent), and distances (5' to 40'). The camera locations are indicated by the red dots, and the camera always points at the sign.}
  \label{fig:angle-setup}
\end{figure}

\begin{table}[tb]
    \centering
    \setlength{\tabcolsep}{6pt}
    \caption{
    Our \textbf{high-confidence} perturbations succeed at attacking at a variety of distances and angles. 
    For each distance-angle combination, we show the detected class and the confidence score. 
    If more than one bounding boxes are detected, we report the highest-scoring one. Confidence values lower than 30\% is considered undetected.
    } 
    \begin{tabular}{rrlrlrlr}
        \toprule
         \textbf{Distance} & \textbf{Angle} & \textbf{person} & (Conf.) & \textbf{sports ball} & (Conf.) & \textbf{untargeted} & (Conf.)\\
        \cmidrule(lr){1-2} \cmidrule(lr){3-4} \cmidrule(lr){5-6} \cmidrule(lr){7-8}
        5' & \ang{0} & person & (.77) & sports ball & (.61) & clock & (.35)\\
        5' & \ang{15} & person & (.91) & cake & (.73) & clock & (.41) \\
        5' & \ang{30} & person & (.93) & cake & (.66) & cake & (.39) \\
        5' & \ang{45} & person & (.69) & cake & (.61) & \textit{stop sign} & (.62) \\
        5' & \ang{60} & \textit{stop sign} & (.93) & \textit{stop sign} & (.70) & \textit{stop sign} & (.88) \\    
        \midrule
        10' & \ang{0} & person & (.55) & cake & (.34) & clock & (.99) \\
        10' & \ang{15} & person & (.63) & cake & (.33) & clock & (.99) \\
        10' & \ang{30} & person & (.51) & cake & (.55) & clock & (.99) \\
        \midrule
        15' & \ang{0} & undetected & --- & cake & (.49) & clock & (.99) \\
        15' & \ang{15} & person & (.57) & cake & (.53) & clock & (.99) \\
        \midrule
        20' & \ang{0} & person & (.49) & sports ball & (.98) & clock & (.99) \\
        20' & \ang{15} & person & (.41) & sports ball & (.96) & clock & (.99) \\
        \midrule
        25' & \ang{0} & person & (.47) & sports ball & (.99) & \textit{stop sign} & (.91) \\
        30' & \ang{0} & person & (.49) & sports ball & (.92) & undetected & --- \\
        40' & \ang{0} & person & (.56) & sports ball & (.30) & \textit{stop sign} & (.30) \\
        \bottomrule
        \multicolumn{2}{r}{Targeted success rate} & 87\% &  & 40\% &  & N/A &  \\
        \midrule
        \multicolumn{2}{r}{Untargeted success rate} & 93\% &  & 93\% &  & 73\% &  \\
        \bottomrule
    \end{tabular}
    \label{tab:angle_high}
\end{table}

\begin{table}[tb]
    \centering
    \setlength{\tabcolsep}{6pt}
    \caption{As expected, low-confidence perturbations achieve lower success rates.} 
    \begin{tabular}{rrlrlrlr}
        \toprule
         \textbf{Distance} & \textbf{Angle} & \textbf{person} & (Conf.) & \textbf{sports ball} & (Conf.) & \textbf{untargeted} & (Conf.)\\
        \cmidrule(lr){1-2} \cmidrule(lr){3-4} \cmidrule(lr){5-6} \cmidrule(lr){7-8}
        5' & \ang{0} & \textit{stop sign} & (.87) & cake & (.90) & cake & (.41)\\
        5' & \ang{15} & \textit{stop sign} & (.63) & cake & (.93) & cake & (.34) \\
        5' & \ang{30} & person & (.83) & cake & (.84) & \textit{stop sign} & (.48) \\
        5' & \ang{45} & \textit{stop sign} & (.97) & \textit{stop sign} & (.94) & \textit{stop sign} & (.82) \\
        5' & \ang{60} & \textit{stop sign} & (.99) & \textit{stop sign} & (.99) & \textit{stop sign} & (.89) \\    
        \midrule
        10' & \ang{0} & \textit{stop sign} & (.83) & \textit{stop sign} & (.99) & undetected & --- \\
        10' & \ang{15} & \textit{stop sign} & (.79) & \textit{stop sign} & (.94) & undetected & --- \\
        10' & \ang{30} & \textit{stop sign} & (.60) & \textit{stop sign} & (.98) & \textit{stop sign} & (.78) \\
        \midrule
        15' & \ang{0} & \textit{stop sign} & (.52) & \textit{stop sign} & (.94) & \textit{stop sign} & (.31) \\
        15' & \ang{15} & \textit{stop sign} & (.33) & \textit{stop sign} & (.93) & undetected & --- \\
        \midrule
        20' & \ang{0} & \textit{stop sign} & (.42) & sports ball & (.73) & undetected & --- \\
        20' & \ang{15} & person & (.51) & sports ball & (.83) & cell phone & (.62) \\
        \midrule
        25' & \ang{0} & \textit{stop sign} & (.94) & sports ball & (.87) & undetected & --- \\
        30' & \ang{0} & \textit{stop sign} & (.94) & sports ball & (.95) & \textit{stop sign} & (.79) \\
        40' & \ang{0} & \textit{stop sign} & (.95) & undetected & --- & \textit{stop sign} & (.52) \\
        \bottomrule
        \multicolumn{2}{r}{Targeted success rate} & 13\% &  & 27\% &  & N/A &  \\
        \midrule
        \multicolumn{2}{r}{Untargeted success rate} & 13\% &  & 53\% &  & 53\% &  \\
        \bottomrule
    \end{tabular}
    \label{tab:angle_low}
\end{table}

\subsubsection{Indoor Experiments}
Following the experimental setup of~\cite{evtimov2017robust}, we took photos of the printed adversarial stop sign, at a variety of distances (5' to 40') and angles (\ang{0}, \ang{15}, \ang{30}, \ang{45}, \ang{60}, from the sign's tangent).
This setup is depicted in \autoref{fig:angle-setup} where camera locations are indicated by red dots.
The camera always pointed at the sign.
We intended these distance-angle combinations to mimic a vehicle's points of view as it would approach the sign from a distance~\cite{lu2017adversarial}.
\autoref{tab:angle_high} and \autoref{tab:angle_low} summarize the results for our \textit{high-confidence} and \textit{low-confidence} perturbations, respectively.
For each distance-angle combination, we show the detected class and the detection's confidence score.
If more than one bounding boxes are detected, we report the highest-scoring one.
Confidence values lower than 30\% were considered undetected; we decided to use the threshold of 30\%, instead of the default 50\% in the Tensorflow Object Detection API~\cite{huang2017speed}, to impose a stricter requirement on ourselves (the ``attacker'').
Since an object can be detected as a stop sign and the target class simultaneously, we consider our attack to be successful only when the confidence score of the target class is the highest among all of the detected classes.

\autoref{tab:angle_high} shows that our high-confidence perturbations achieve a high attack success rate at a variety of distances and angles.
For example, we achieved a targeted success rate 87\% in misleading the object detector into detecting the stop sign as a \textit{person}, and an even higher untargeted success rate of 93\% when our attack goal is to cause the detector to either fail to detect the stop sign (e.g., at 15' \ang{0}) or to detect it as a class that is \textit{not} a stop sign. The \textit{sports ball} targeted attack has a lower targeted success rate but achieves the same untargeted success rate.  Our untargeted attack consistently misleads the detection into the \textit{clock} class in medium distances, but is less robust for longer distances.
%
Overall, the perturbation is less robust to very high viewing angle (\ang{60} from the sign's tangent), because we did not simulate the viewing angle distortion in the optimization.

The low-confidence perturbations (\autoref{tab:angle_low}), as expected, achieve a much lower attack success rate, suggesting the need to use higher-confidence perturbations when conducting the more challenging drive-by tests 
(described in the next section).
\autoref{tab:indoors_samples} shows some sample high-confidence perturbations from our indoor experiments.

\begin{table}[tb]
\centering
\setlength{\tabcolsep}{4pt}
\caption{Sample high-confidence perturbations from indoor experiments. For complete experiment results, please refer to \autoref{tab:angle_high}.}
\begin{tabular}{rrm{1.2in}m{1.2in}m{1.2in}}

\toprule
\textbf{Dist.} & \textbf{Angle} & \textbf{Target: person} & \textbf{Target: sports ball} & \textbf{Untargeted}\\
\midrule
40' & \ang{0} 
& \includegraphics[width=1.2in]{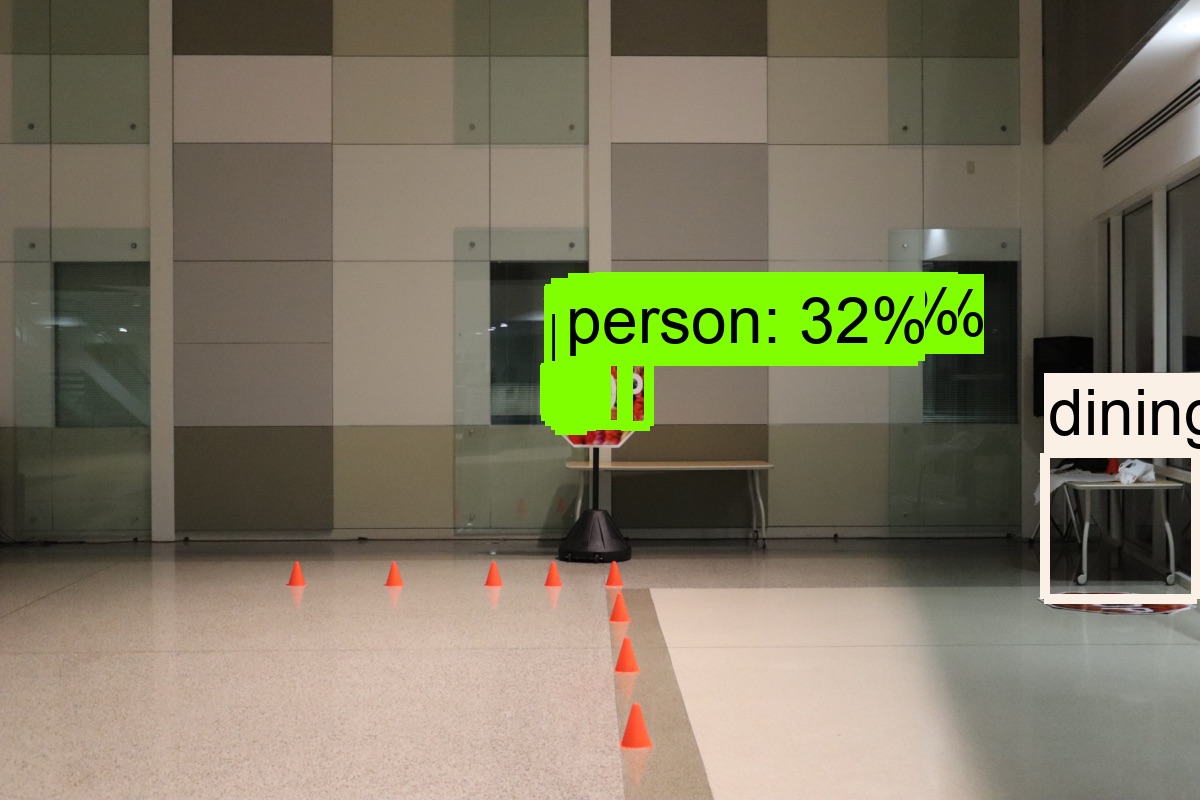} 
& \includegraphics[width=1.2in]{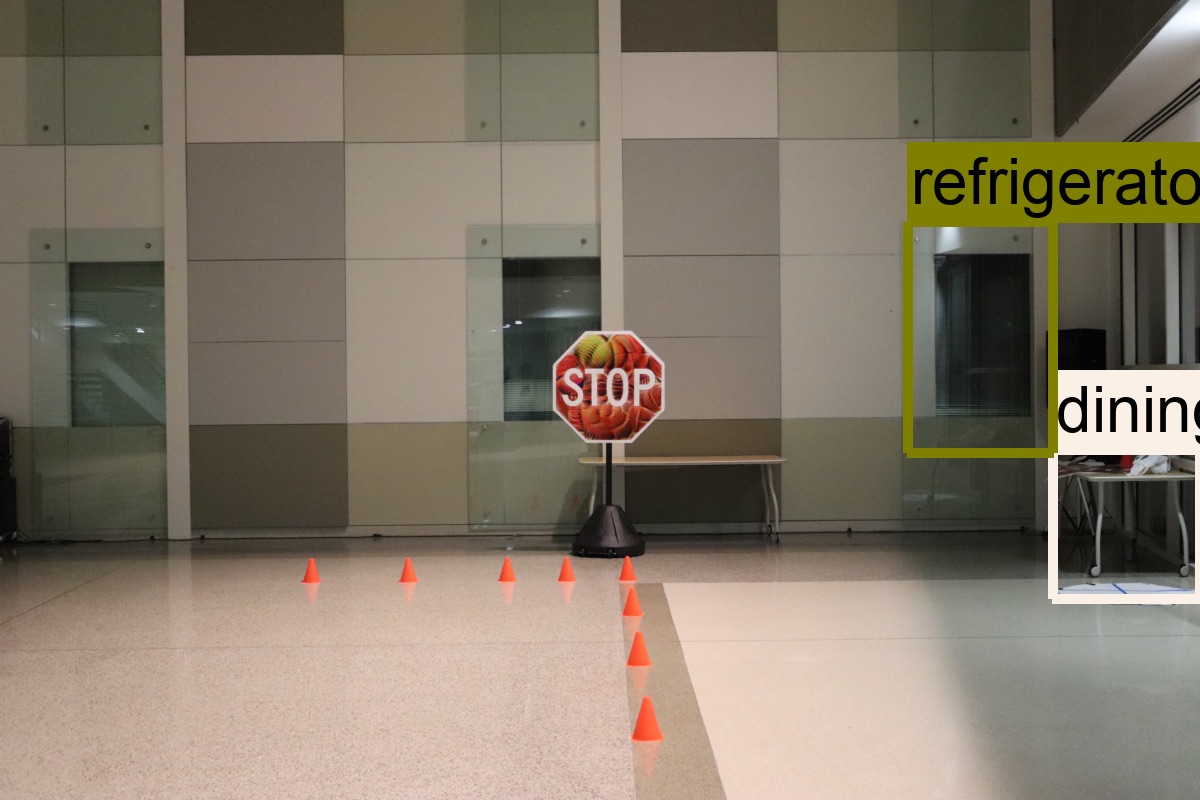} 
& \includegraphics[width=1.2in]{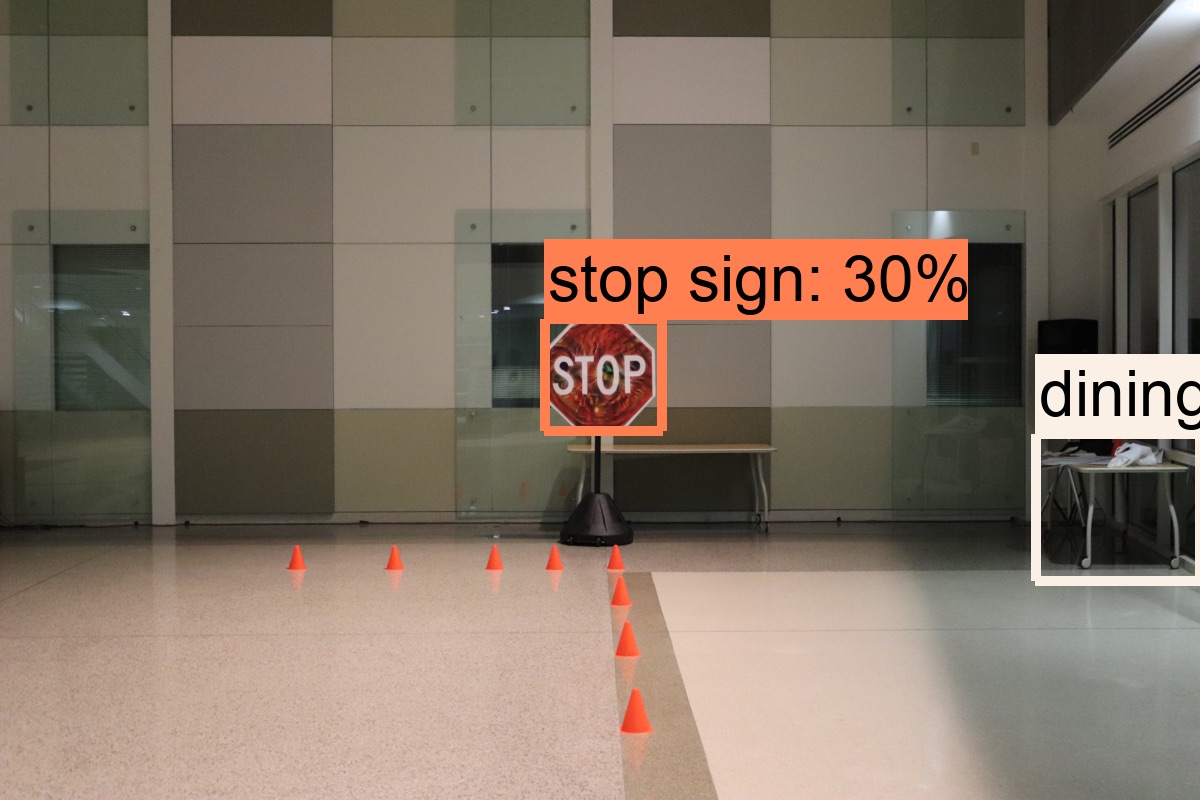}\\
10' & \ang{0} 
& \includegraphics[width=1.2in]{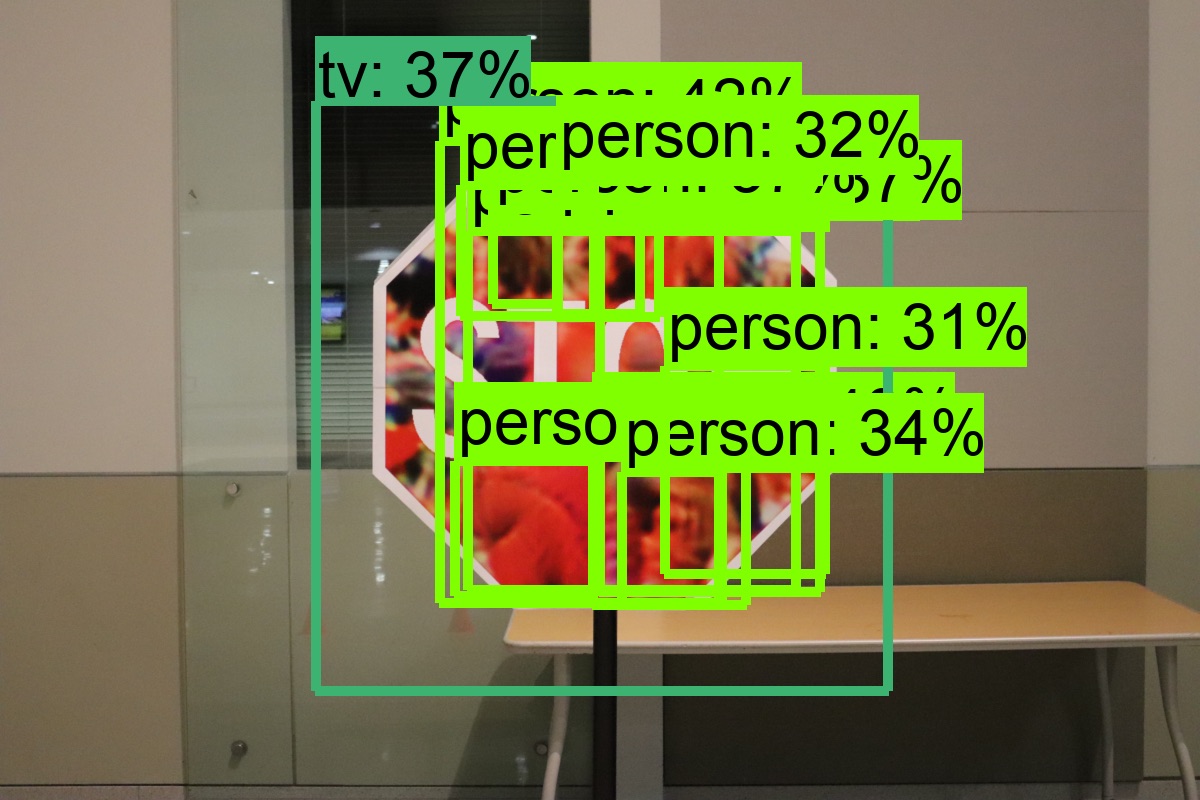} 
& \includegraphics[width=1.2in]{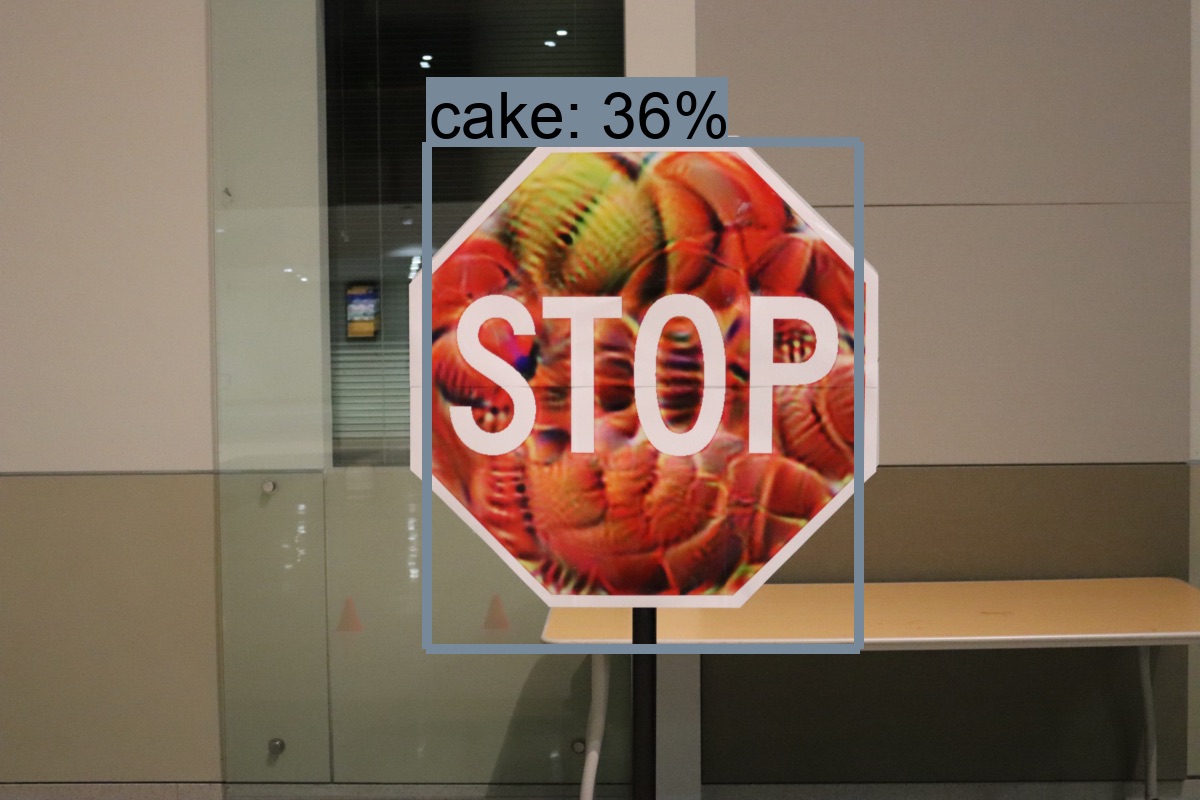} 
& \includegraphics[width=1.2in]{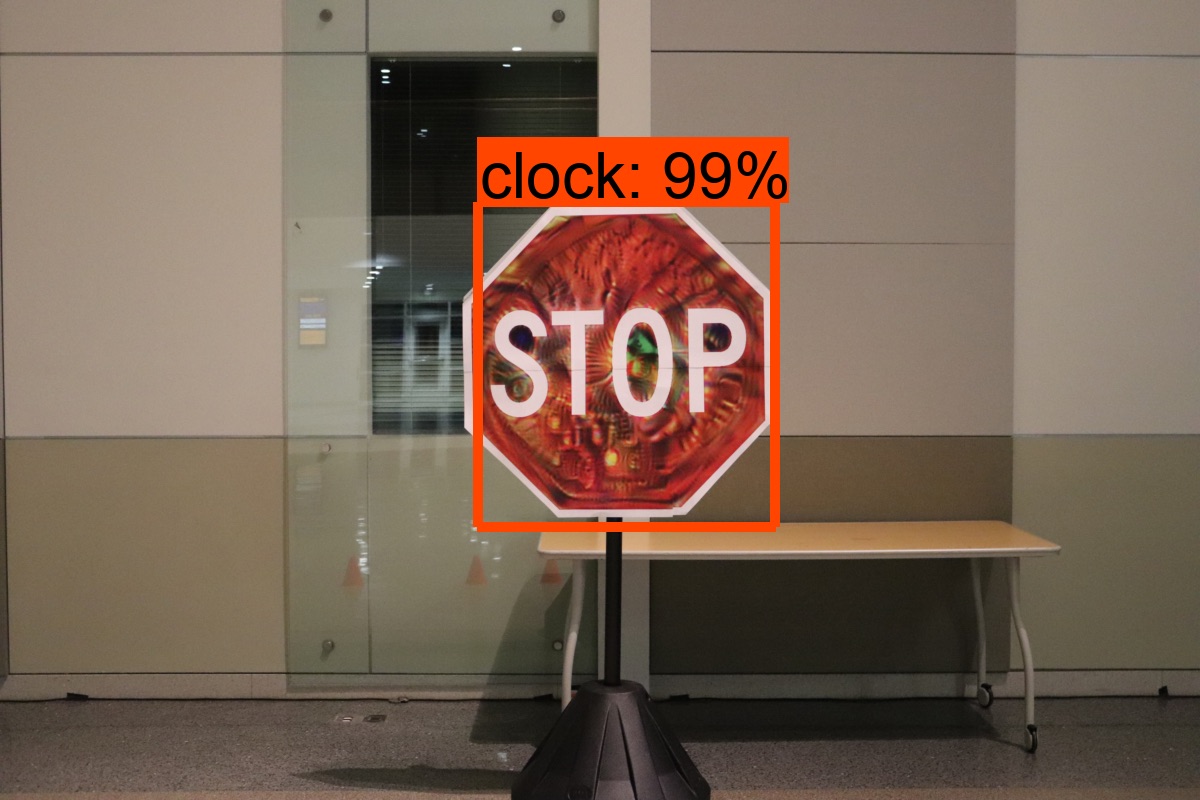}\\
10' & \ang{30} 
& \includegraphics[width=1.2in]{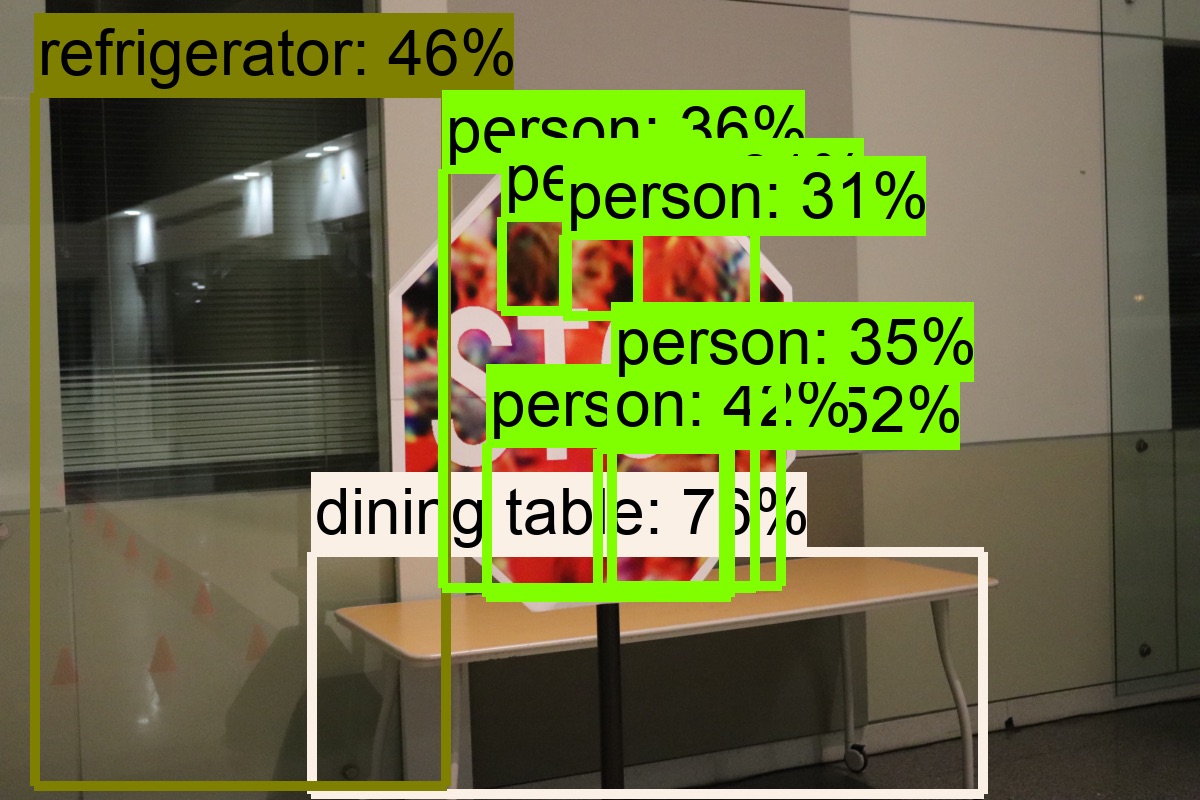} 
& \includegraphics[width=1.2in]{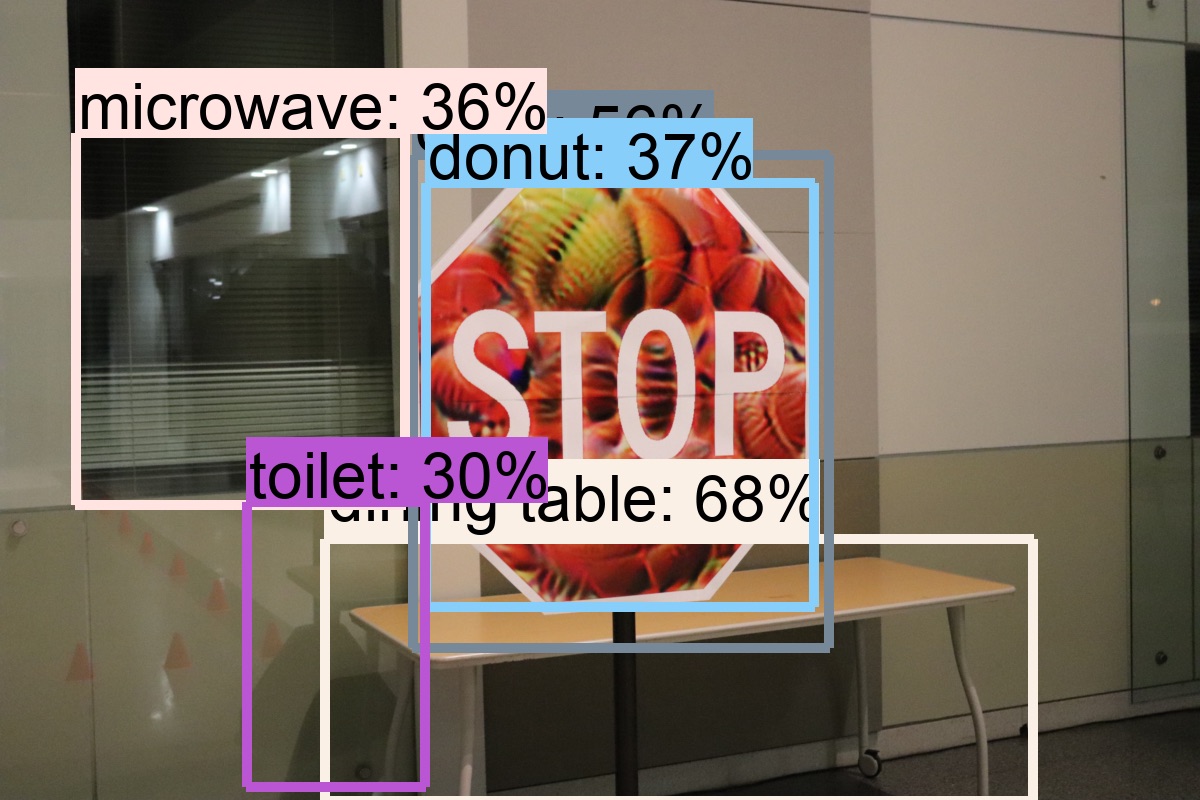} 
& \includegraphics[width=1.2in]{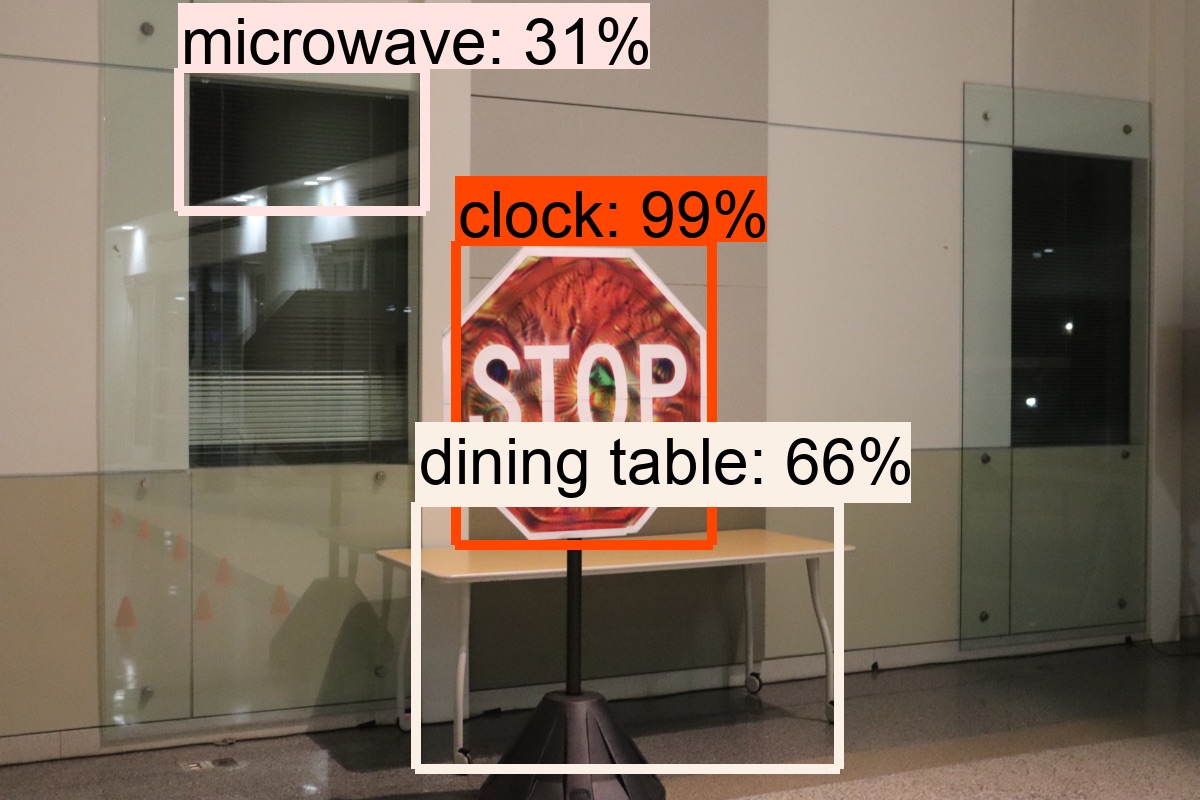}\\
5' & \ang{60} 
& \includegraphics[width=1.2in]{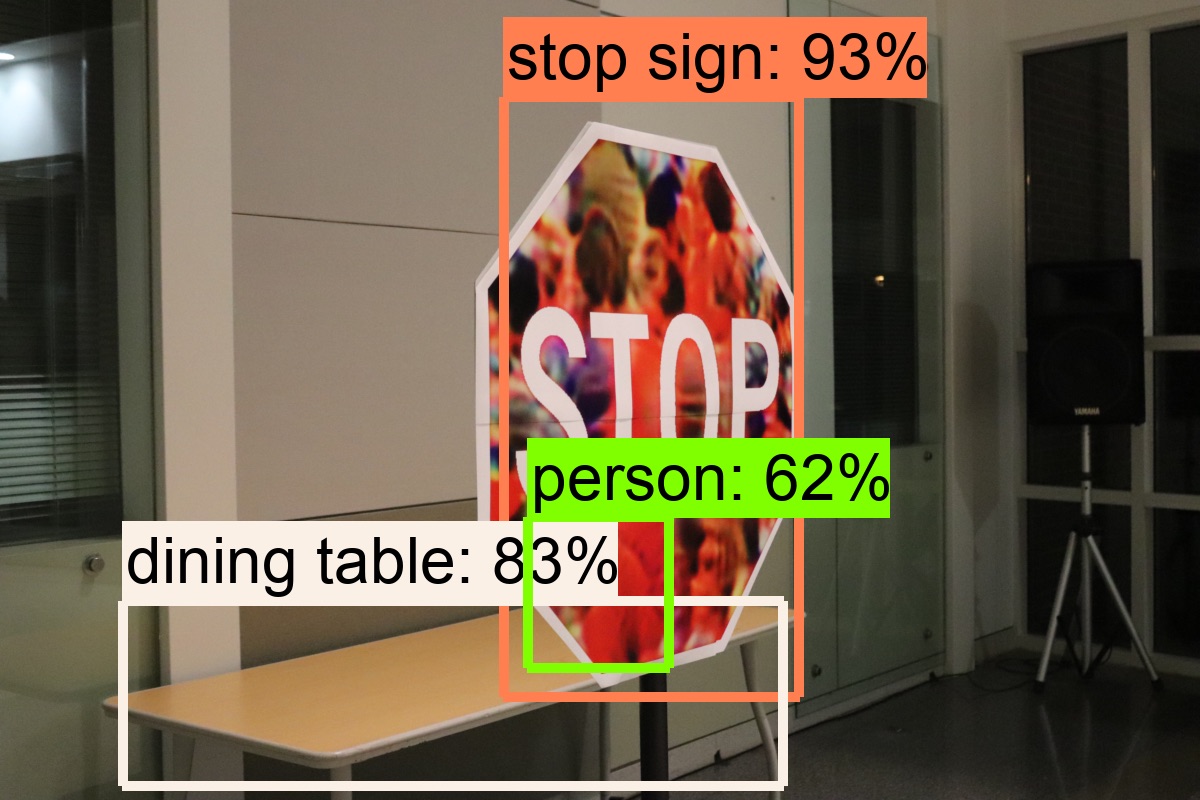} 
& \includegraphics[width=1.2in]{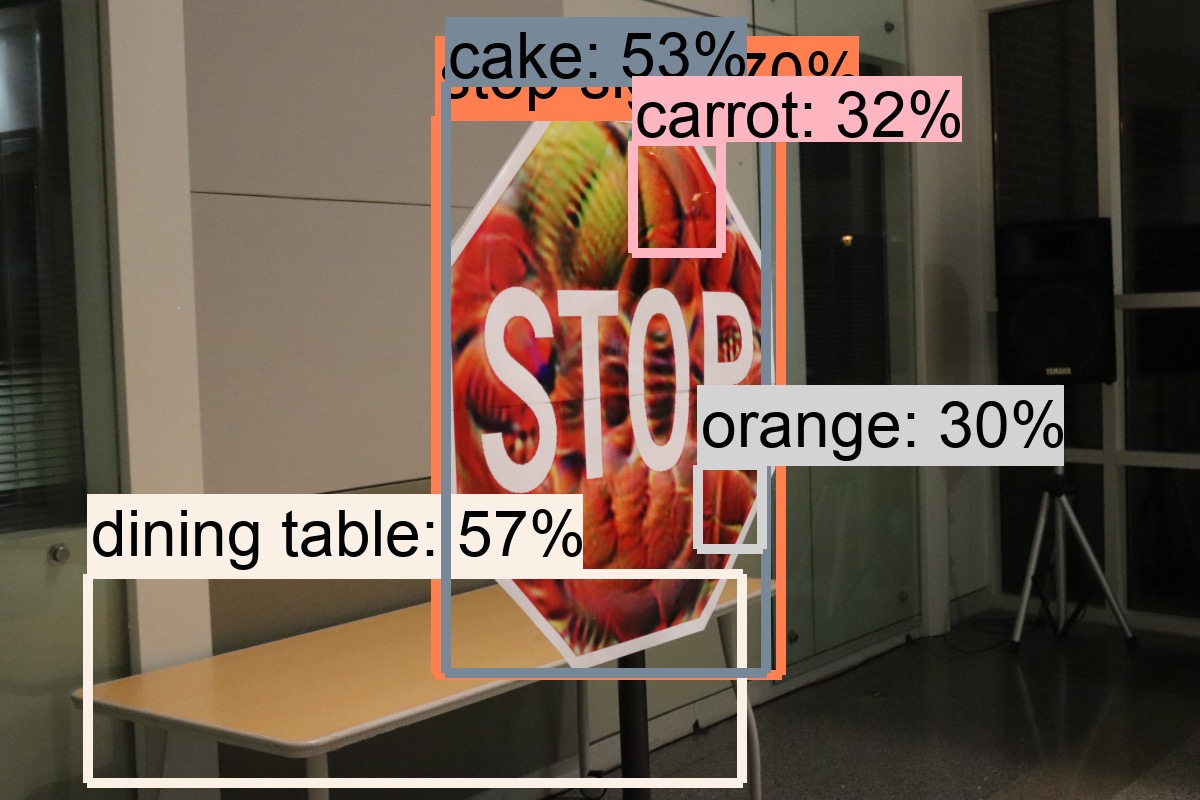} 
& \includegraphics[width=1.2in]{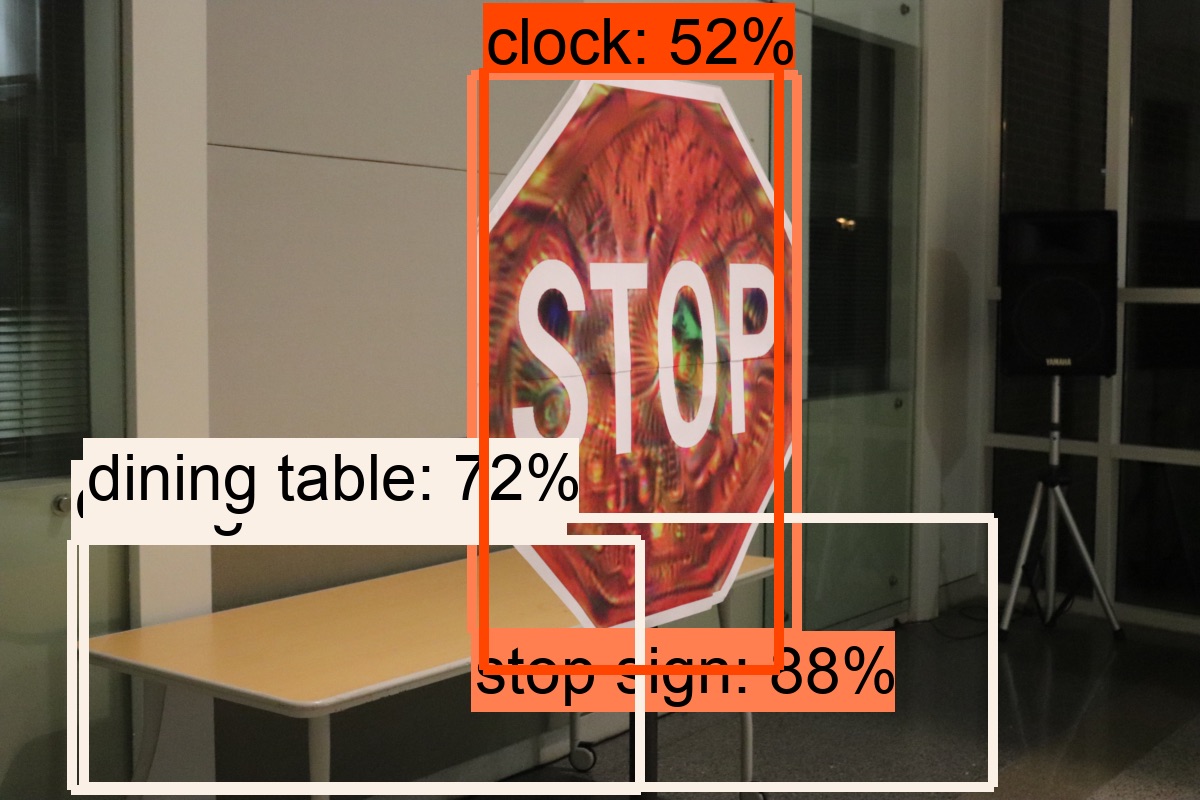}\\
\bottomrule

\end{tabular}
\label{tab:indoors_samples}
\end{table}


\subsubsection{Drive-by Tests}
We performed drive-by tests at a parking lot so as not to disrupt other vehicles with our stop signs.
We put a purchased real stop sign as a control and our printed perturbed stop sign side by side.
Starting from about 200 feet away, we slowly drove (between 5~mph to 15~mph) towards the signs while simultaneously recording video from the vehicle's dashboard at 4K resolution and 24 FPS using an iPhone~8 Plus.
We extracted all video frames, and for each frame, we obtained the detection results from Faster R-CNN object detection model.
Because our low confidence attacks showed relatively little robustness indoors, we only include the results from our high-confidence attack.
Similar to our indoor experiments, we only consider detections that had a confidence score of at least 30\%.

\begin{figure}[h!]
  \centering
  \includegraphics[width=\textwidth]{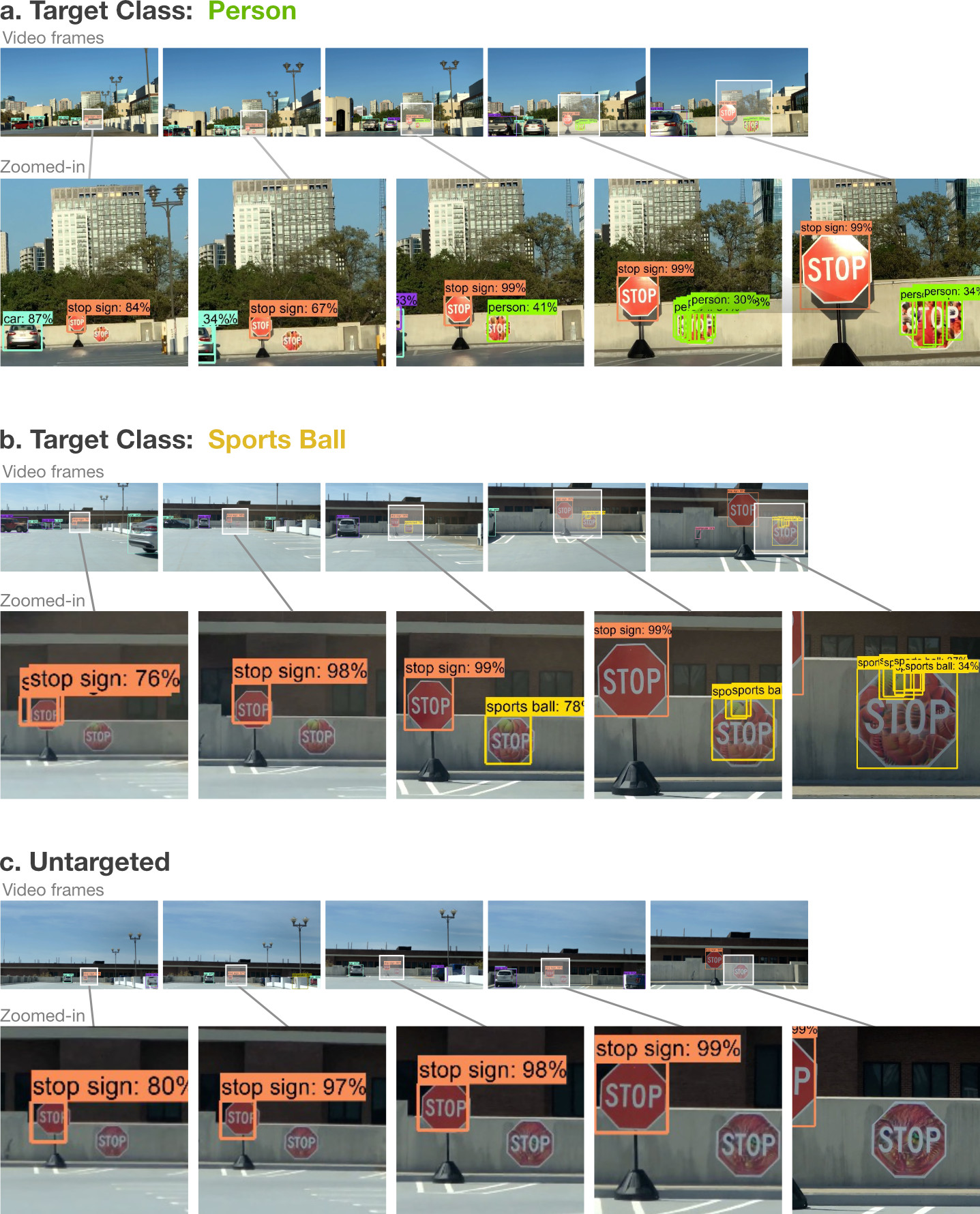}
  \caption{Snapshots of the drive-by test results. In (a), the person perturbation was detected 47\% of the frames as a person and only once as a stop sign. The perturbation in (b) was detected 36\% of the time as a sports ball and never as a stop sign. The untargeted perturbation in (c) was detected as \textit{bird} 6 times and never detected as a stop sign or anything else for the remaining frames.}
  \label{fig:drive_by_snapshots}
\end{figure}

In \autoref{fig:drive_by_snapshots}, we show sample video frames (rectangular images) to give the readers a sense of the size of the signs relative to the full video frame; 
we also show zoomed-in views (square images) that more clearly show the Faster R-CNN detection results.

The \textit{person-perturbation} in \autoref{fig:drive_by_snapshots}a drive-by totaled 405 frames as partially shown in the figure.
The real stop sign in the video was correctly detected in every frame with high confidence.
On the other hand, the perturbed stop sign was only correctly detected once, while 190 of the frames identified the perturbed stop sign as a person with medium confidence.
For the rest of the 214 frames the object detector failed to detect anything around the perturbed stop sign.

The video we took with the \textit{sports-ball-perturbation} shown in \autoref{fig:drive_by_snapshots}b had 445 frames.
The real stop sign was correctly identified all of the time, while the perturbed stop sign was never detected as a stop sign.
As the vehicle (video camera) moved closer to the perturbed stop sign, 160 of the frames were detected as a sports ball with medium confidence. One frame was detected as \textit{apple} and \textit{sports ball} and the remaining 284 frames had no detection around the perturbed stop sign.

Finally, the video of the untargeted perturbation (\autoref{fig:drive_by_snapshots}c) totaled 367 frames.
While the unperturbed stop sign was correctly detected all of the time, the perturbed stop sign was detected as \textit{bird} 6 times and never detected for the remaining 361 frames.

\subsubsection{Exploring Black-box Transferability}
We also sought to understand how well our high-confidence perturbations could fool other object detection models.
For image recognition, it is known that high-confidence targeted attacks fail to transfer~\cite{liu2017delving}.

To this end, we fed our high-confidence perturbations into 8 other MS-COCO-trained models from the Tensorflow detection model zoo\footnote{\url{https://github.com/tensorflow/models/blob/master/research/object_detection/g3doc/detection_model_zoo.md}}.
Table~\ref{tab:transferability} shows how well our perturbation generated from the Faster R-CNN Inception-V2 transfer to other models.
To better understand transferability, we examined the worse case.
That is, if a model successfully detects a stop sign in the image, we say the perturbation has failed to transfer or attack that model.
We report the number of images (of the 15 angle-distance images in our indoor experiments) where a model successfully detected a stop sign with at least 30\% confidence.
We also report the maximum confidence of all of those detected stop sign.

\begin{table}[tb]
    \centering
    \setlength{\tabcolsep}{3pt}
    \caption{Black-box transferability of our 3 perturbations. We report the number of images (of the 15 angle-distance images) that failed to transfer to the specified model. We consider the detection of any stop sign a ``failure to transfer.'' Our perturbations fail to transfer for most models, most likely due to the iterative nature of our attack.} 
    \begin{tabular}{lrrrrrr}
        \toprule
         \textbf{Model} & \textbf{person} & (conf.) & \textbf{sports ball} & (conf.) & \textbf{untargeted} & (conf.) \\
        \cmidrule(lr){1-1} \cmidrule(lr){2-3} \cmidrule(lr){4-5} \cmidrule(lr){6-7}
        Faster R-CNN Inception-V2 & 3  & (.93) & 1  & (.70) & 5  & (0.91) \\
        \midrule
        SSD MobileNet-V2          & 2  & (.69) & 8  & (.96) & 15 & (1.00)\\
        SSD Inception-V2          & 11 & (1.00) & 14 & (.99) & 15 & (1.00)\\
        R-FCN ResNet-101          & 4  & (.82) & 10 & (.85) & 15 & (1.00)\\
        Faster R-CNN ResNet-50    & 13 & (.00) & 15 & (1.00) & 15 & (1.00)\\
        Faster R-CNN ResNet-101   & 15 & (.99) & 13 & (.97) & 15 & (1.00)\\
        Faster R-CNN Inc-Res-V2   & 1  & (.70) & 0  & (.00) & 12 & (1.00)\\
        Faster R-CNN NASNet       & 14 & (1.00) & 15 & (1.00) & 15 & (1.00)\\
        \bottomrule
    \end{tabular}
    \label{tab:transferability}
\end{table}

Table~\ref{tab:transferability} shows the lack of transferability of our generated perturbations.
The untargeted perturbation fails to transfer most of the time, followed by the sports ball perturbation, and finally the person perturbation.
The models most susceptible to transferability were the Faster R-CNN Inception-ResNet-V2 model, followed by the SSD MobileNet-V2 model.
Iterative attacks on image recognition also usually fail to transfer~\cite{liu2017delving}, so it is not surprising that our attacks fail to transfer as well.
We leave the thorough exploration of transferability as future work.

\section{Discussion \& Future Work}
\begin{figure}[h!]
  \centering
  \includegraphics[width=\textwidth]{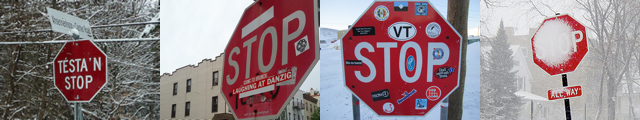}
  \caption{Example stop signs from the MS-COCO dataset.
  Stop signs can vary by language, by degree of occlusion by stickers or modification by graffiti, or just elements of the weather. Each stop sign in the images is correctly detected by the object detector with high confidence (99\%, 99\%, 99\%, and 64\%, respectively).}
  \label{fig:coco_examples}
\end{figure}

There is considerable variation in the physical world that real systems will have to deal with.
\autoref{fig:coco_examples} shows a curated set of non-standard examples of stop signs from the MS-COCO dataset\footnote{Full resolution images of the examples in \autoref{fig:coco_examples} can be found at: \url{http://cocodataset.org/\#explore?id=315605}, \url{http://cocodataset.org/\#explore?id=214450}, \url{http://cocodataset.org/\#explore?id=547465}, and \url{http://cocodataset.org/\#explore?id=559484}}.
The examples show stop signs in a different language, or that have graffiti or stickers applied to them, or that have been occluded by the elements.
In each of these cases, it is very unlikely a human would misinterpret the sign as anything else but a stop sign.
They each have the characteristic octagonal shape and are predominantly red in color.
Yet, the object detector sees something else.

Unlike previous work on adversarial examples for image recognition, our adversarial perturbations are overt.
They, like the examples in \autoref{fig:coco_examples}, exhibit large deviations from the standard stop sign.
A human would probably notice these large deviations, and a trained human might even guess they were constructed to be adversarial.
But they probably would not be fooled by our perturbations.
However an automated-system using an off-the-shelf object detector would be fooled, as our results show.
Our digital perturbation shown in \autoref{fig:ball_high} does look like a baseball or tennis ball has been painted on the upper right hand corner.
\autoref{fig:drive_by_snapshots}b shows how the object detector detects this part of the image as a sports ball with high confidence.
This might seem unfair, but attackers have much latitude when these kind of models are deployed in automated systems.
Even in non-automated systems a human might not think anything of \autoref{fig:person_high} because it does not exhibit any recognizable person-like features.

Attackers might also generate perturbations without restricting the shape and color, and attach
them to some arbitrary objects, like a street light or a trash bin. An untrained eye might see these perturbations as some kind of artwork, but the autonomous system might see something completely different. This attack, as described in~\cite{sitawarin2018darts}, could be extended to object detectors using our method.

Defending against these adversarial examples has proven difficult.
Many defenses fall prey to the so-called ``gradient masking'' or ``gradient obfuscating'' problem~\cite{athalye2018obfuscated}.
The most promising defense, adversarial training, has yet to scale up to models with good performance on the ImageNet dataset. Whether adversarial training can mitigate our style of overt, large-deviation (e.g., large $\ell_p$ distance) perturbations is also unclear.

\section{Conclusion}
We show that the state-of-the-art Faster R-CNN object detector, while previously considered more robust to physical adversarial attacks, can actually be attacked with high confidence.
Our work demonstrates vulnerability in MS-COCO-learned object detectors and posits that security and safety critical systems need to account for the potential threat of adversarial inputs to object detection systems.

Many real-world systems probably do not use an off-the-shelf pre-trained object detector as in our work.
Why would a system with safety or security implications care to detecting sports balls?
Most probably do not.
Although it remains to be shown whether our style of attack can be applied to safety or security critical systems that leverage object detectors, our attack provides the means to test for this new class of vulnerability.

{\bf Acknowledgements:}
This work is supported in part by NSF grants CNS-1704701, TWC-1526254, and a gift from Intel.

\bibliographystyle{splncs04}
\bibliography{main}

\end{document}